\documentclass[sigconf]{acmart}
\usepackage{subcaption}
\usepackage{tabularx}

\AtBeginDocument{%
  }

\begin{document}

\title{Evolutionary Level Repair}

\author{Debosmita Bhaumik}
\affiliation{%
  \institution{Institute of Digital Games}
  \city{Msida}
  \country{Malta}
}
\email{debosmita.bhaumik.22@um.edu.mt}

\author{Julian Togelius}
\affiliation{%
  \institution{Game Innovation Lab}
  \city{New York}
  \state{New York}
  \country{USA}
}
\email{julian@togelius.com}

\author{Georgios N. Yannakakis}
\affiliation{%
  \institution{Institute of Digital Games}
  \city{Msida}
  \country{Malta}
}
\email{georgios.yannakakis@um.edu.mt}

\author{Ahmed Khalifa}
\affiliation{%
  \institution{Institute of Digital Games}
  \city{Msida}
  \country{Malta}
}
\email{ahmed@akhalifa.com}

\renewcommand{\shortauthors}{Bhaumik et al.}

\begin{abstract}
We address the problem of game level repair, which consists of taking a designed but non-functional game level and making it functional. This might consist of ensuring the completeness of the level, reachability of objects, or other performance characteristics. The repair problem may also be constrained in that it can only make a small number of changes to the level. We investigate search-based solutions to the level repair problem, particularly using evolutionary and quality-diversity algorithms, with good results. This level repair method is applied to levels generated using a machine learning-based procedural content generation (PCGML) method that generates stylistically appropriate but frequently broken levels. This combination of PCGML for generation and search-based methods for repair shows great promise as a hybrid procedural content generation (PCG) method.
\end{abstract}



\received{20 February 2007}
\received[revised]{12 March 2009}
\received[accepted]{5 June 2009}

\maketitle

\section{Introduction}
Many types of game content, such as levels, exhibit a dual nature where both visual aesthetics and functionality are important. The way the content looks is important, not only for the visual appeal but also for what it communicates about how it can be played. But functionality is, in many games, even more crucial; it determines what is possible to do in the game and what leads to success and failure. For example, a Super Mario Bros level gives a sense of thematic unity, foreshadows challenges, and gently tells the player where to go. Still, it also defines the challenges that need to be overcome. The fusion of visual aesthetics and functionality is central to what a game level is, but functionality is more important. While a playable level that looks like a mess is still passable game content, a beautiful but non-completable level is game-breaking.

Training (or fine-tuning) a self-supervised learning model of the type usually used for visual data can therefore yield a model that outputs pretty but broken levels. For example, training an autoencoder on a set of levels from the classic puzzle platformer Lode Runner\cite{bhaumik2021lode} results in a neural network that generates nice-looking levels that mostly cannot be solved. After training GANs to generate Mario levels, most of the naively sampled content is not functional~\cite{volz2018evolving}. Many machine learning models can not, on their own, ensure functionality.

So, how can you generate content that both looks good and plays well? One approach is to train a model on existing levels using self-supervised methods, and then use some kind of search algorithm to search the learned space for good content. This is the general approach of latent variable evolution ~\cite{bontrager2018deepmasterprints,volz2018evolving} and latent variable illumination~\cite{fontaine2021illuminating}. Still, this approach introduces certain constraints on the machine learning method, notably that it has a latent space, which might exclude some ML methods. Even after using these methods and restricting the latent search around the broken level, there is still a chance that the space might not contain a repaired version of it. This reduces the learned latent space to a smaller subset and discards a huge part of the search space because of small mistakes in it, which causes the method to have less creative freedom.

Another approach is to generate and then repair. The problem of level repair then becomes one of making sure the level functions correctly or follows some defined constraints (e.g., that it is playable and contains the appropriate challenges) while maintaining the overall visual aesthetics of the level. Preserving the visual aesthetics can intuitively be achieved by limiting the number of changes made. Different work has combined constraint solving~\cite{bazzaz2024guided,cooper2022sturgeon}, linear programming~\cite{zhang2020video}, machine learning models~\cite{jain2016autoencoders,chen2020image}, and search-based methods~\cite{liapis2013sentient,cooper2020pathfinding} with PCGML to battle the functionality issue. Inspired by the ``generate then repair'' we are trying to combine the advantage of machine learning methods with evolutionary methods to repair unplayable levels. 

\begin{figure}
    \centering
    \begin{subfigure}[t]{0.45\linewidth}
        \centering
        \includegraphics[width=\linewidth]{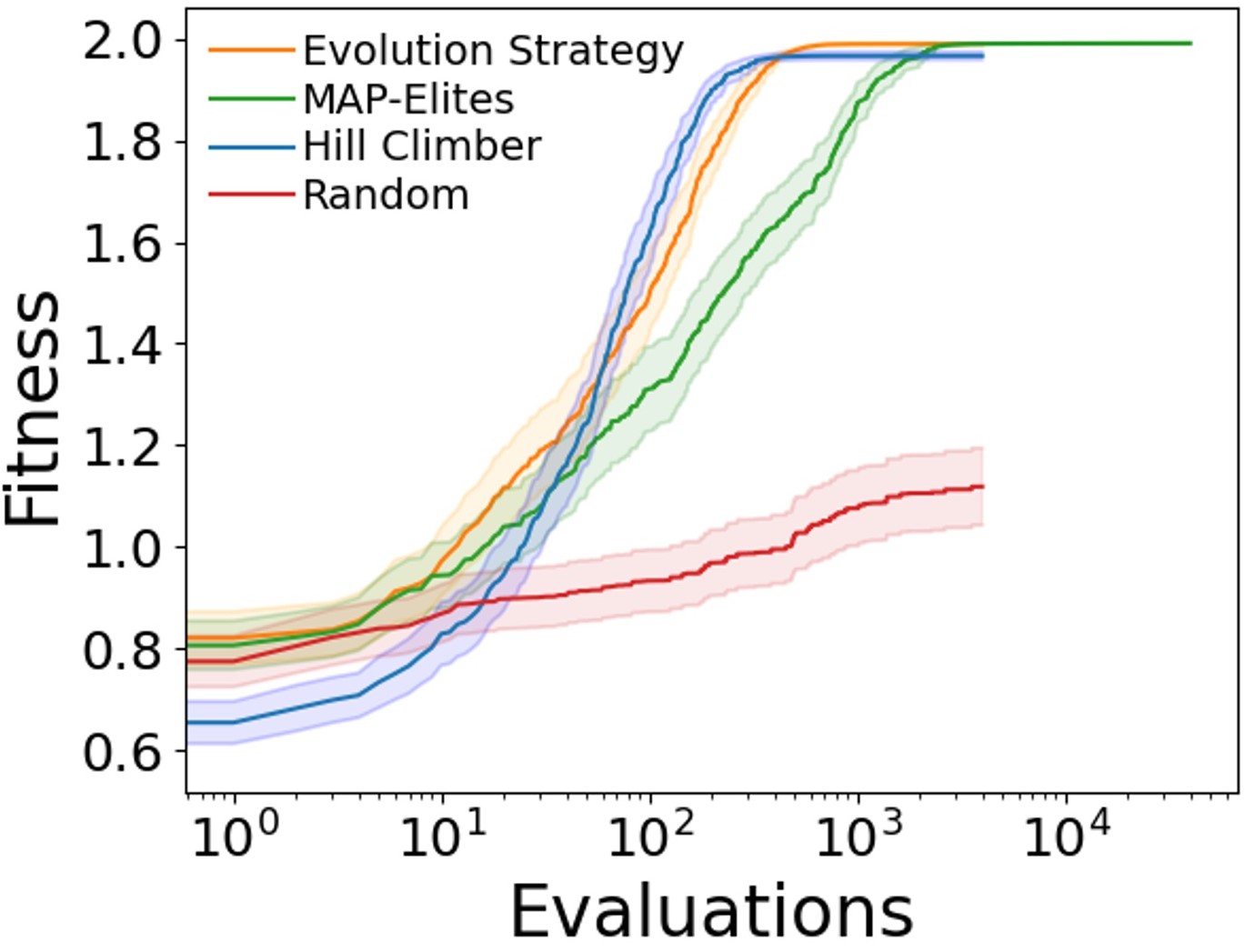}
        \caption{Maximum Fitness during optimization.}
        \label{fig:fitness}
    \end{subfigure}
    \begin{subfigure}[t]{0.45\linewidth}
        \centering
        \includegraphics[width=\linewidth]{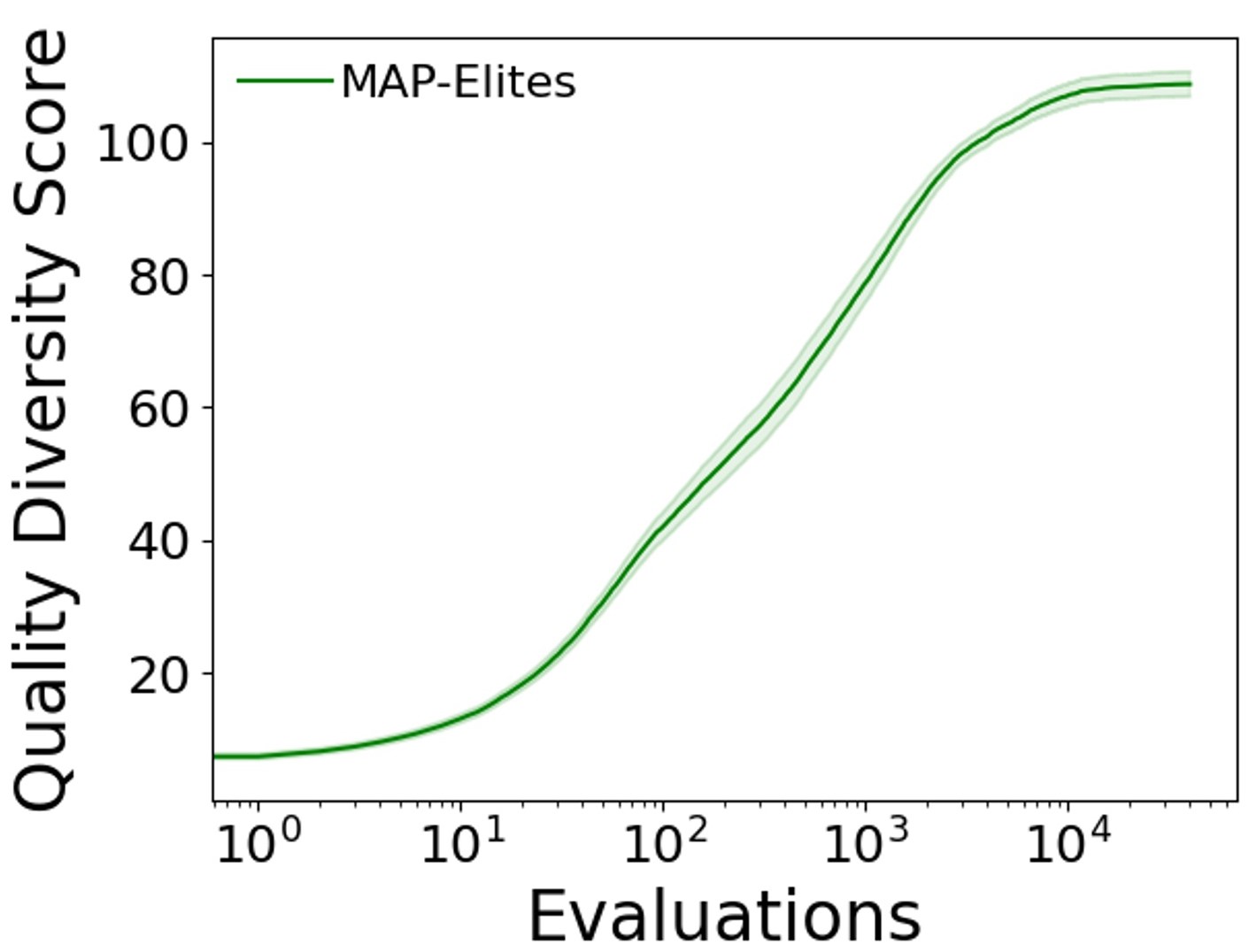}
        \caption{Quality diversity score of the MAP-Elites algorithm.}
        \label{fig:elites}
    \end{subfigure}
    \caption{The average performance for all four algorithms with 95\% confidence interval over 200 runs.}
    \label{fig:performance}
\end{figure}

\section{Methods}
Repairing broken game levels is not a straightforward task, especially when the goal is to make the level functional. A good repair function will try to make a small number of changes to the existing content to make it playable. If the repaired level differs significantly from the visual aesthetic of the given level, it is generally not a good repair. At the same time, a good evaluation metric is required to measure the changes made by the repair function and guide the process in the right direction. For repairs, we use two different algorithms: $\mu + \lambda$ Evolution Strategy (ES)~\cite{beyer2007evolution} for normal optimization and MAP-Elites (ME)~\cite{mouret2015illuminating} for Quality Diversity.

We have selected Lode Runner for this project. Lode Runner levels are of size $32x22$ tiles and made of eight different tile types: empty, brick, solid-brick, rope, ladder, gold, enemy, and player. The objective of the game is to collect all the gold in the level without being caught by the enemies. The player can move horizontally on brick tiles and ropes and navigate vertically using ladders. Lode Runner was selected as our domain problem due to its relatively large level, which makes it easy to see when a level looks ``noisy''.

\subsection{Representation and Operators}

We use direct representation, where the levels are represented as a 1D array of tiles (flattened representation of the $32x22$ level). For modifying the chromosomes, we only use mutation. The mutation operator picks $1$ to $m$ (m is the maximum number of modifications at each step) random locations and changes them. The tile values are changed either randomly by selecting any other tile or copying the corresponding tile from the starting level for that location. The probability of which mutation will be applied is determined based on the fitness value. Lower fitness favors random mutation to explore more towards the beginning of the evolution, while higher fitness encourages copy mutation to ensure that the repaired level still looks similar to the starting level. One important note on the mutation is that we removed the ability to erase or add gold tiles or the player tile to force evolution to repair the level by making it connected instead of removing unreachable gold. Finally, the initial population is created by applying random mutations on the starting level.

\subsection{Fitness Function and Characteristics}
In our problem, the fitness function needs to deal with playability and similarity. Both are equally important for a good repair function, also difficult to balance. To improve playability, tiles need to be modified, which affects the similarity score in the opposite direction. To solve this, we are using a \emph{cascading elitism}~\cite{togelius2007towards} fitness function. This encourages evolution to prioritize playability first, then it tries to improve the similarity as shown in equation~\ref{eq:total}.
\begin{equation}\label{eq:total}
    f_{total} = \begin{cases}
        f_{playability} & \text{if $gold_{collect} < gold_{total}$}\\
        1 + f_{similarity} & \text{otherwise}
    \end{cases}
\end{equation}
where $gold_{total}$ is the total number of golds and $gold_{collect}$ is the number of reachable golds. 

\begin{table}
    \begin{tabularx}{\columnwidth}{c|c|c|c}
        \hline
        Starting Playability  & 30\% - 50\%  & 50\% - 70\%  & 70\% - 90\% \\
        \hline
        Random Search & 8.8 ± 1.34 & 10.72 ± 1.72 & 2.51 ± 1.13 \\
        Hill Climber & 10.35 ± 1.02 & 21.43 ± 11.37 & 35.05 ± 11.39 \\
        Evolution Strategy & 9.21 ± 1.02 & 5.4 ± 0.94 & 4.66 ± 0.65 \\
        MAP-Elites & \textbf{7.91 ± 0.68} & \textbf{4.63 ± 0.67} & \textbf{4.22 ± 0.53} \\
        \hline
    \end{tabularx}
    \caption{Average number of changes applied by each algorithm with 95\% confidence interval to repair.}
    \label{tab:repair_changes}
    \vspace{-20pt}
\end{table}

The playability score consists of the ratio of reachable golds in the level and the exploration score by the flood-fill algorithm, as shown in equation~\ref{eq:playability}. The exploration score is added to help smooth the fitness function. The number of reachable golds is calculated using a flood-fill algorithm from the player's starting location. This algorithm expands only using the possible player's actions at each tile except for digging (for simplicity).
\begin{equation}\label{eq:playability}
    f_{playability} = min(1, \frac{gold_{collect} + \frac{tiles_{explored}}{lvl_{size}}}{gold_{total}})
\end{equation}
where $gold_{total}$ is the total number of golds, $gold_{collect}$ is the number of reachable golds, $tiles_{explored}$ is the number of tiles explored by the flood-fill agent, and $lvl_{size}$ is the size of the level.

For similarity, we use the hamming distance between the current level and the starting level as shown in equation~\ref{eq:similarity}.
\begin{equation}\label{eq:similarity}
    f_{similarity} = \frac{lvl_{size}-dist(lvl, lvl_{start})}{lvl_{size}}
\end{equation}
where $lvl$ is the current level, $lvl_{start}$ is the starting level, $lvl_{size}$ is Lode Runner level size (22x32), and $dist(lvl, lvl_{start})$ is the hamming distance between the starting level and the current level.

Behavior characteristics are features that can be calculated for each chromosome. Behavior characteristics help the quality diversity algorithm to find solutions that are different from each other. Since, the connectivity is what we care about towards playability, the different ways to reach this connectivity are important to differentiate between solutions. We used 2 simple behavior characteristics that correlate with connectivity: the number of added Ropes and the Number of added Ladders. The ladder tile is the only tile that allows the player to walk in all 4 directions, and the rope tile allows them to walk in 3 different directions (no up).

\begin{figure}[!tb]
    \centering
    \includegraphics[width=\linewidth]{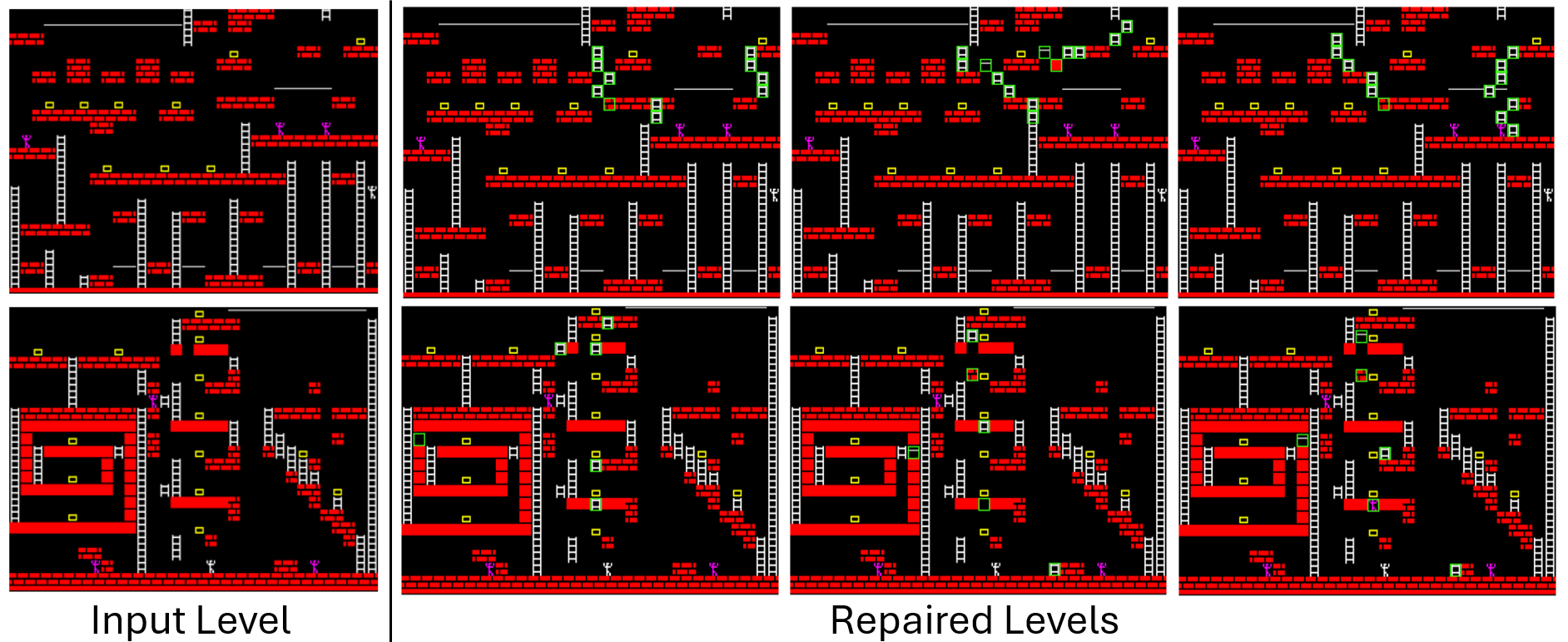}
    \caption{Example of $\mu + \lambda$ evolution strategy repaired levels.}
    \label{fig:repair_evol}
\end{figure}

\section{Experiments}

For our experiments, we used 20 unplayable levels, 10 were generated using the Wave Function Collapse algorithm (WFC)~\cite{karth2017wavefunctioncollapse} and the other 10 levels using a trained autoencoder~\citet{bhaumik2021lode}. We place the player on a randomly selected empty tile, as a good Lode Runner level should be fully connected, meaning that a player can reach all gold tiles from any location. 

While generating unplayable levels from WFC and the autoencoder, we discarded any level that had less than 30\% playability (the player cannot collect more than 30\% of the gold) as repairing these levels might need too many modifications. We wanted to explore the performance of the proposed repair methods on different types of broken levels. We divided the starting levels into three groups based on their starting playability score ($30-50\%$, $50-70\%$, and $70-90\%$).

We have applied the ES and the ME algorithm on each level, and repeated them 10 times to showcase the difference and stability between runs. We ran both evolutions until fitness stopped improving (figure~\ref{fig:fitness}). For the ES, this was 4,000 generations with $\mu = \lambda = 50$, which is 200k evaluations. For ME, this was 2 million evaluations with $100$ chromosomes as a start. The number of mutations ($m$) is limited to a maximum of 10 to allow small changes in the levels. The probability of random mutation and copying tile mutation is linearly sampled between $80/20\%$ at fitness of $0$ to $20/80\%$ at fitness of $2$. Finally, each behavior characteristic is divided into 9 bins [<0, 0, 5, 10, 15, 20, 60, 100, 100+]. This division is based on some initial experiments where we find that this is the best balance between the number of bins and finding a good solution fast.

We applied a random search algorithm and a hill climber (HC)~\cite{russell2016artificial} as baseline algorithms. HC starts with a broken level and at each step selects a random location from the level, then picks the best possible modification at that location. On the other hand, the random search (RS) generates multiple new levels by applying random changes to the starting level (0-20 tiles) and keeps the best-found solution. Both HC and RS ran for 200k evaluations, which is the same number of evaluations as ES. 

\section{Results}

\begin{figure*}
    \centering
    \begin{subfigure}[t]{0.49\linewidth}
        \centering
        \includegraphics[width=\linewidth]{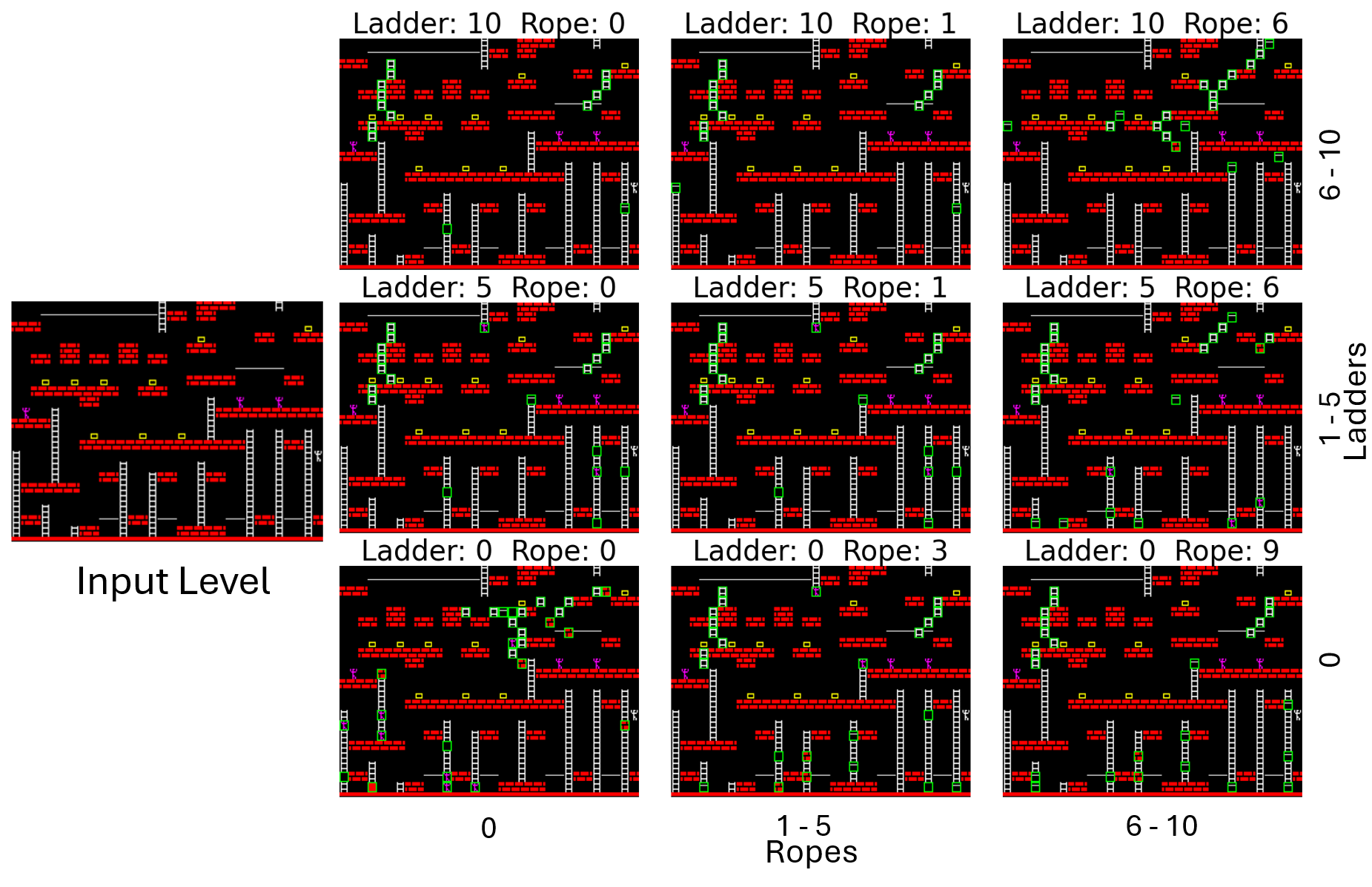}
    \end{subfigure}
    \begin{subfigure}[t]{0.49\linewidth}
        \centering
        \includegraphics[width=\linewidth]{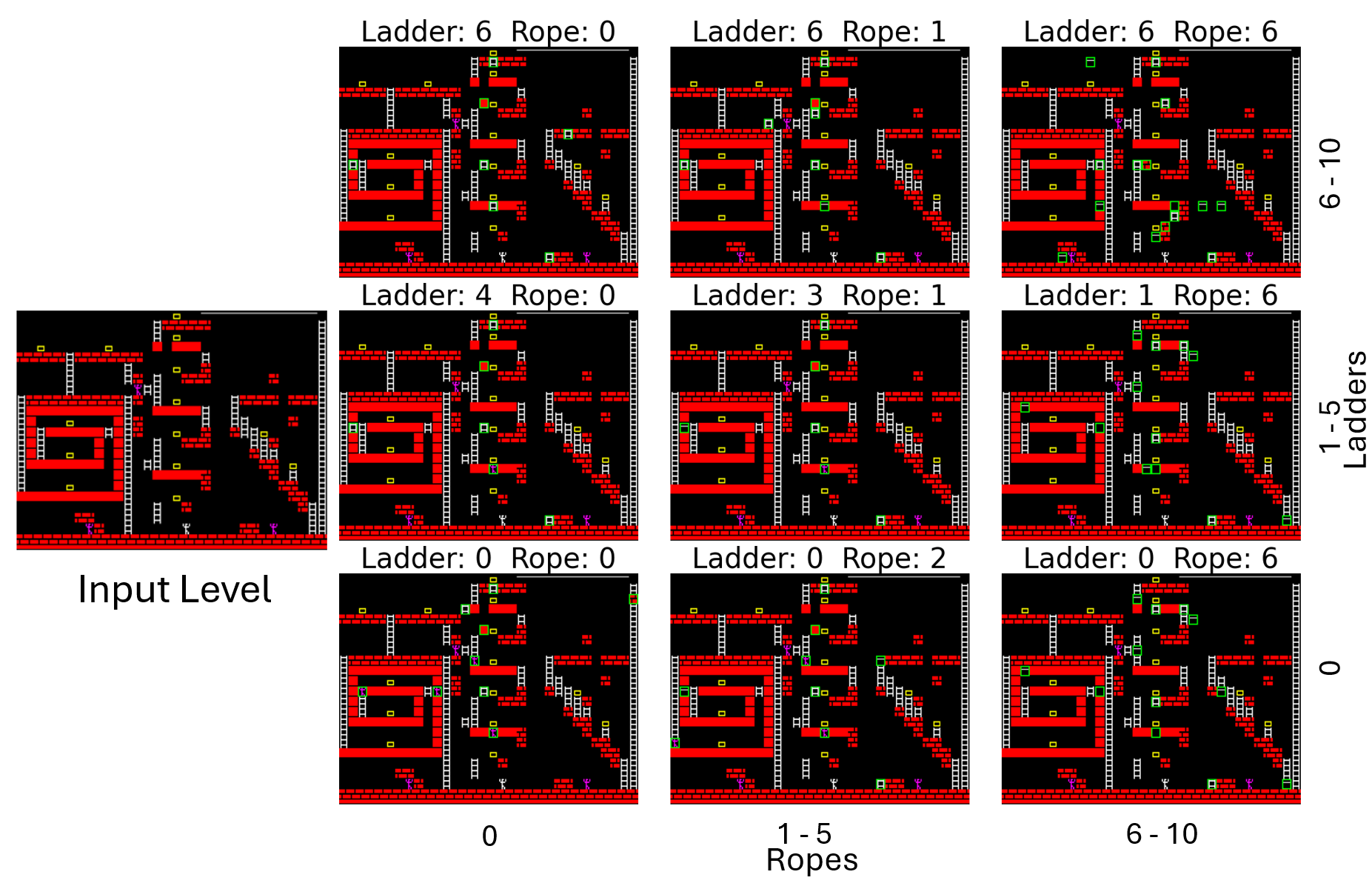}
    \end{subfigure}
    \caption{The repaired levels from the bottom corner of the MAP-Elites archive (Ropes and Ladders as behavior characteristic).}
    \label{fig:repair_me}
\end{figure*}

HC, ES, and ME managed to repair all input levels in different time scales, while the RS succeeded only 26\% of the time. Figure~\ref{fig:fitness} shows the average of max fitness between all the runs with the 95\% confidence interval for different algorithms. It shows the ES and ME reached a little higher fitness compared to the HC in 200k evaluations. We believe that the local nature of HC prevents it from exploring more optimal repairs (that require fewer changes). We see that ME took a little longer to reach the highest fitness and then stabilized. To understand why, we show the quality diversity (QD) score (sum of the fitness of all solutions in the archive) of the ME archive with the 95\% confidence interval in figure~\ref{fig:elites}. We can see that the QD score starts stabilizing around 2 million evaluations. This is reasonable as the MAP-Elites tried to improve the entire archive instead of finding a single solution compared to ES or HC (as shown in figure~\ref{fig:repair_me}). 

\begin{figure}
    \centering
    \begin{subfigure}[t]{0.49\linewidth}
        \centering
        \includegraphics[width=\linewidth]{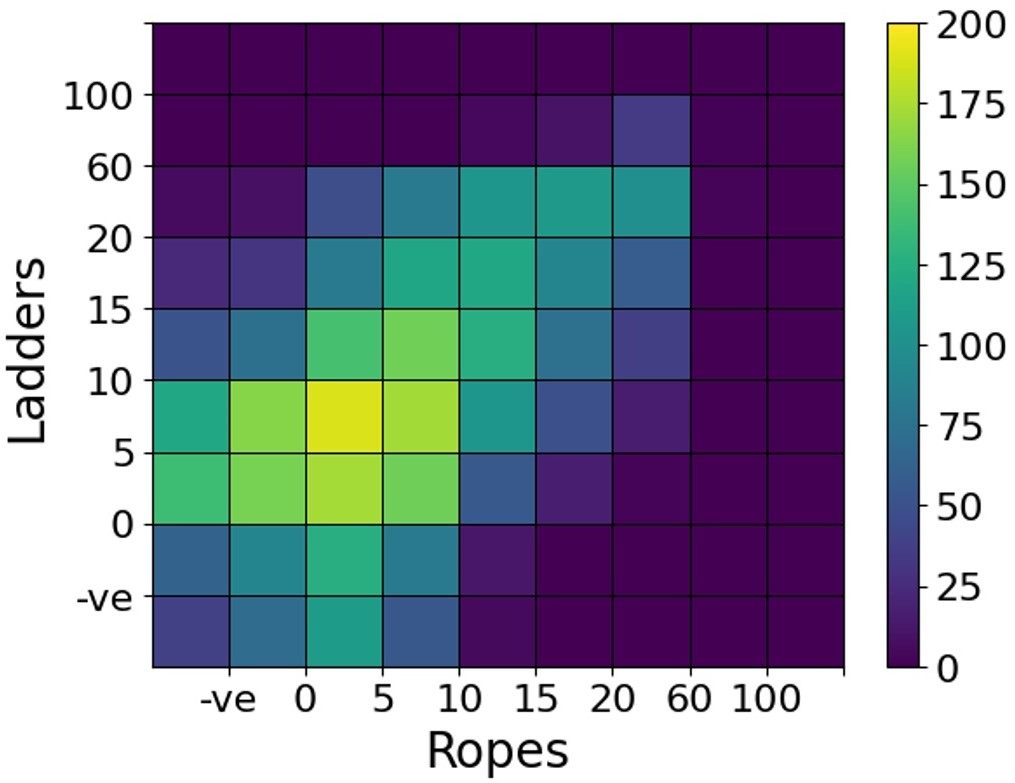}
        \caption{Evolution Strategy}
        \label{fig:arc_es}
    \end{subfigure}
    \begin{subfigure}[t]{0.49\linewidth}
        \centering
        \includegraphics[width=\linewidth]{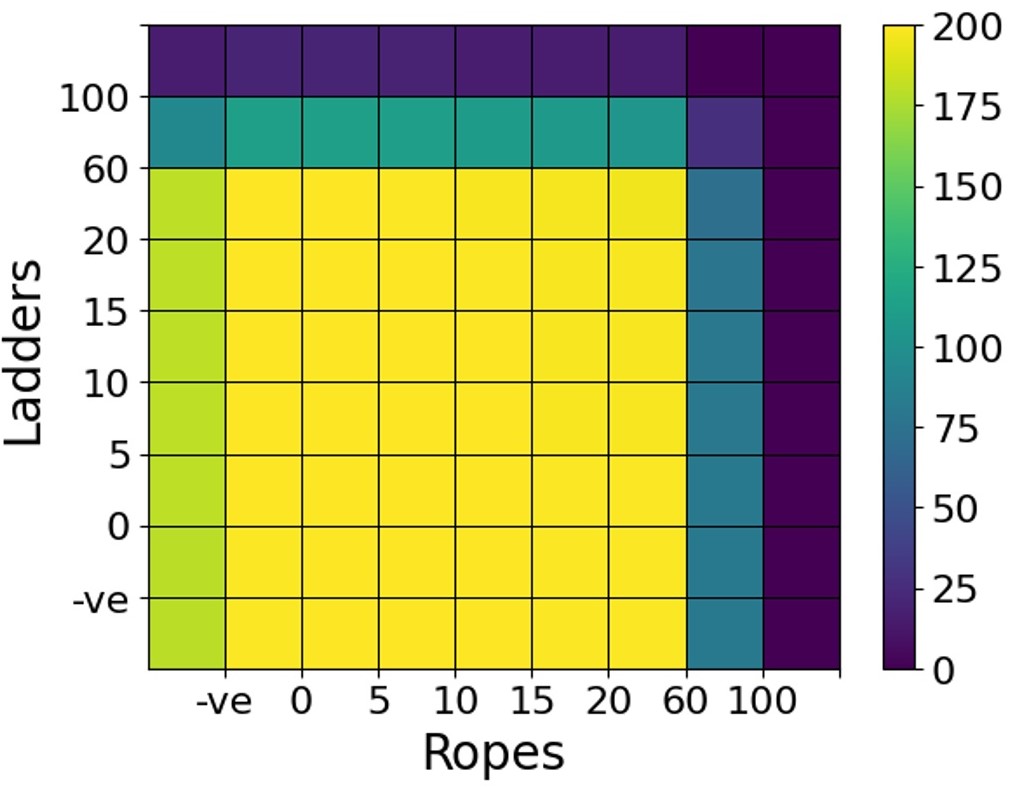}
        \caption{MAP-Elites}
        \label{fig:arc_me}
    \end{subfigure}
    \caption{Heatmap of all the discovered solutions over all the runs by $\mu + \lambda$ Evolution Strategy and MAP-Elites.}
    \label{fig:arc_heatmap}
    \vspace{-10pt}
\end{figure}

To understand more about the repairs, we compared all the algorithms based on the number of changes in each starting level group. Table~\ref{tab:repair_changes} shows the average number of changes applied on levels with 95\% confidence interval. The HC applies higher changes for all groups compared to the others, with the highest values for $50-70\%$ and $70-90\%$ groups. We believe that some of these levels might have more than one way to repair, where the longer repair path has a better fitness landscape increase compared to others. When comparing the changes applied by ME and the ES, for levels with the lowest playability($30-50\%$), the number of changes is maximum, as these levels have a higher number of unreachable golds and to connect them, big changes are required. For the other two groups they are quite close to each other, with ME performing a bit better. We think the behavior characteristics helped the ME to explore different areas of the search space and discover better repairs overall. This can also be seen in figure~\ref{fig:arc_heatmap} where the ME algorithm covers the low change cells more compared to the ES. 

Figure~\ref{fig:repair_evol} shows the levels repaired from different runs of the ES. It is visible that the evolution strategy made very few changes to make the level playable. One noticeable attribute is that most repairs involve adding extra ladders, as ladders allow movement in all 4 directions compared to other tiles, but this limits the diversity of the repaired levels as they show similar ways of traversing the levels. Another observation is that the added ladders are placed diagonally instead of a straight line. This is due to the fitness that tries to minimize the amount of tiles modified; diagonal connections allow for more area with fewer changes.

Figure~\ref{fig:repair_me} displays the repairs done by the ME algorithm. We show only the bottom corner of the archive (without <0 cells) as we are only interested in solutions that involve a small number of changes. Out of 10 different runs, we display the one with the highest quality diversity score. One interesting feature in these repairs can be seen in the 0 bins of the behavior characteristic. As the algorithm discovers that they can't be solved without adding ladders so the evolution decided to reuse ladders from less needed areas. This is done by removing the ladder from a specific area and adding it to another. Finally, sometimes when the repair does not require extra ladders or ropes, the algorithm just adds random ones to cover that cell in the archive. This problem can be solved by changing the behavior characteristics to only show used ladders and ropes.

To understand the diversity of solutions coming from ES compared to ME, we save playable levels from each generation of ES in an archive with a similar structure to the ME archive (figure~\ref{fig:arc_heatmap}). We compared the number of archive cells covered by both algorithms over all the runs. On average, the ES managed to fill $25\%$ of the archive cells (with $5\%$ confidence interval) whereas MAP-Elites covered $68\%$ (with $8\%$ confidence interval) of cells in the archive. This was expected as the ES tried to repair levels using minimum changes without caring about the tiles used. Another observation is that the ES focused on the diagonal parts of the archive with most cells in the lower left area, except for the 0 and negative bins. This was expected as the ES tries to reduce the number of changes from the original level.

\section{Discussion \& Conclusion}

We showed that evolutionary algorithms can be used to efficiently repair broken levels. The proposed methods were compared with two baseline methods, random search and hill climber. The hill climber managed to achieve a close fitness to our proposed methods, but it tends to use a large number of changes (due to their local search nature). On the other hand, the random baseline failed to repair levels 74\% of the time. Evolution Strategy was able to find solutions in 10 times fewer evaluations than MAP-Elites, but with very little control over how the repairs are happening. Having control over what type of repairs is considered better for mixed-initiative tools~\cite{yannakakis2014mixed}. Finally, the MAP-Elites algorithm overall manages to do better repairs with fewer changes than the Evolution Strategy; it showcases the ability of diversity in helping to find better repairs. Perhaps the biggest concern for the method proposed in this paper is the time complexity. Evolutionary PCG is not the fastest method, particularly if a non-trivial evaluation function is used. This raises the question of whether we could train a machine learning method that imitates the repair trajectories discovered here, similar to the work done by Khalifa et al.~\cite{khalifa2022mutation}. That would make level repairing more viable for real-time use cases such as design assistance.

\bibliographystyle{acmart}
\bibliography{biblo}

\end{document}